\pdfoutput=1

\documentclass[11pt]{article}

\usepackage{authblk}
\usepackage[]{acl}

\usepackage{times}
\usepackage{latexsym}

\usepackage[T1]{fontenc}

\usepackage[utf8]{inputenc}

\usepackage{microtype}
\usepackage{graphicx}

\title{A Rationale-Centric Framework for Human-in-the-loop Machine Learning}

\author{
    Jinghui Lu$^*$ \textsuperscript{\rm1,4},
    Linyi Yang$^*$ \textsuperscript{\rm2,3},
    Brian Mac Namee \textsuperscript{\rm1}, 
    Yue Zhang \textsuperscript{\rm2,3} 
    \\
    \textsuperscript{1} The Insight Centre for Data Analytics, University College Dublin \\
    \textsuperscript{2} School of Engineering, Westlake University \\
    \textsuperscript{3} Institute of Advanced Technology, Westlake Institute for Advanced Study \\
    \textsuperscript{4} SenseTime Research \\
    \texttt{\{jinghui.lu, brian.macnamee\}@ucd.ie} \\
    \texttt{\{yanglinyi, zhangyue\}@westlake.edu.cn}\\
    \texttt{}\\
    
}
\date{}

\begin{document}
\maketitle
\def\thefootnote{*}\footnotetext{ These authors contributed equally to this work.}
\begin{abstract}
We present a novel rationale-centric framework with human-in-the-loop -- \textbf{R}ationales-centric \textbf{D}ouble-robustness \textbf{L}earning (RDL) -- to boost model out-of-distribution performance in few-shot learning scenarios. By using static semi-factual generation and dynamic human-intervened correction, RDL exploits rationales (i.e. phrases that cause the prediction), human interventions and semi-factual augmentations to decouple spurious associations and bias models towards generally applicable underlying distributions, which enables fast and accurate generalisation. Experimental results show that RDL leads to significant prediction benefits on both in-distribution and out-of-distribution tests compared to many state-of-the-art benchmarks---especially for few-shot learning scenarios. We also perform extensive ablation studies to support in-depth analyses of each component in our framework.\looseness=-1


\end{abstract}

\section{Introduction}

\begin{figure}[!t]
\centering
\includegraphics[width=.35\textwidth]{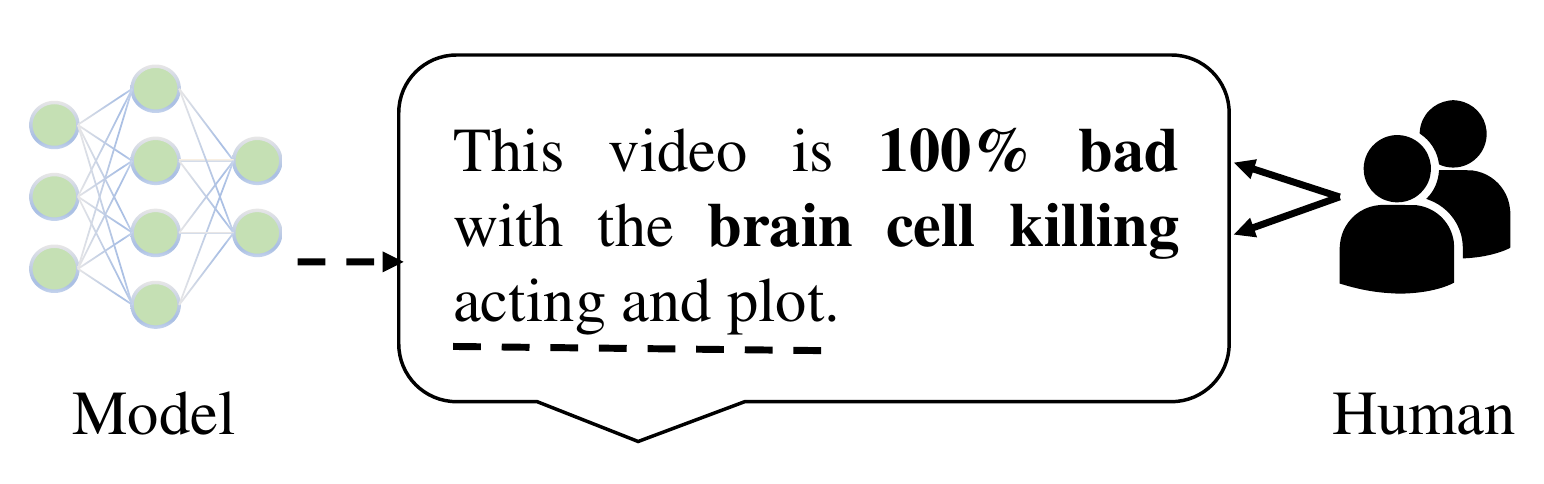}
 \caption{A negative movie review with human annotated causal terms (bold text) and spurious patterns recognised by the model (underlined text).}\label{fig:example}
\end{figure}

Recent work finds that natural artefacts \citep{gururangan2018annotation} or spurious patterns \citep{keith2020text,srivastava2020robustness} in datasets can cause sub-optimal model performance for neural networks. As shown in Figure \ref{fig:example}, the bold phrases---\emph{``\textbf{100\% bad}''} and \emph{``\textbf{brain cell killing}''}---are underlying causes for a negative sentiment prediction that most human readers would recognise. These are defined as \emph{rationales} in this paper. The underlined phrase---``\underline{acting and plot}''--- has been incorrectly recognised as a causal term by the model used fort this example, and is referred to as a \emph{spurious pattern}.\looseness=-1


Spurious patterns (or associations) are caused by natural artefacts or biases in training data \citep{lertvittayakumjorn2021explanation}, and are usually useless, or even harmful, at test time. This issue can be severe in few-shot learning (FSL) scenarios. For instance, \citet{kulesza2010} suggests that when a model is trained with a small subset of labelled data, it is prone to exploiting spurious patterns leading to poor generalisability that is evident in the performance decay in out-of-distribution (OOD) datasets. In spite of these issues, training deep neural networks using few labelled examples is a compelling scenario since unlabelled data may be abundant but labelled data is expensive to obtain in real-world applications \citep{lu2020investigating,10.1007/978-3-030-88942-5_18}.\looseness=-1

There is a strand of research addressing this scenario that seeks to improve model performance by \emph{``introducing methods and resources for training models less sensitive to spurious patterns''} \citep{kaushik2020learning}. Most of this work relies on generating counterfactual augmented data (CAD), either manually \citep{kaushik2021learning} or automatically \citep{feng-etal-2021-empowering,qian2021counterfactual,yang-etal-2021-exploring,yang2020generating,delaney2021uncertainty}. For example, \citet{kaushik2020learning} proposed a human-in-the-loop framework where human annotators are required to make minimal changes to original movie reviews to produce sentiment-flipped counterfactual reviews, which enables models to learn useful associations between input texts and output labels \citep{kaushik2021learning}. 


Generating manual counterfactuals, however, is expensive and time-consuming---\citet{kaushik2020learning} report the cost of revising $2.5k$ instances at over \$10,000. On the other hand, fully automatic methods are task-specific and therefore have weak robustness across domains and less reliability compared to manual counterfactuals. To address these issues, we propose \textbf{R}ationales-centric \textbf{D}ouble-robustness \textbf{L}earning (RDL), a human-in-the-loop framework for data augmentation in a few-shot setting, which is efficient, robust, model-agnostic, and general across tasks.



\begin{figure}[t]
\centering
\includegraphics[width=.5\textwidth]{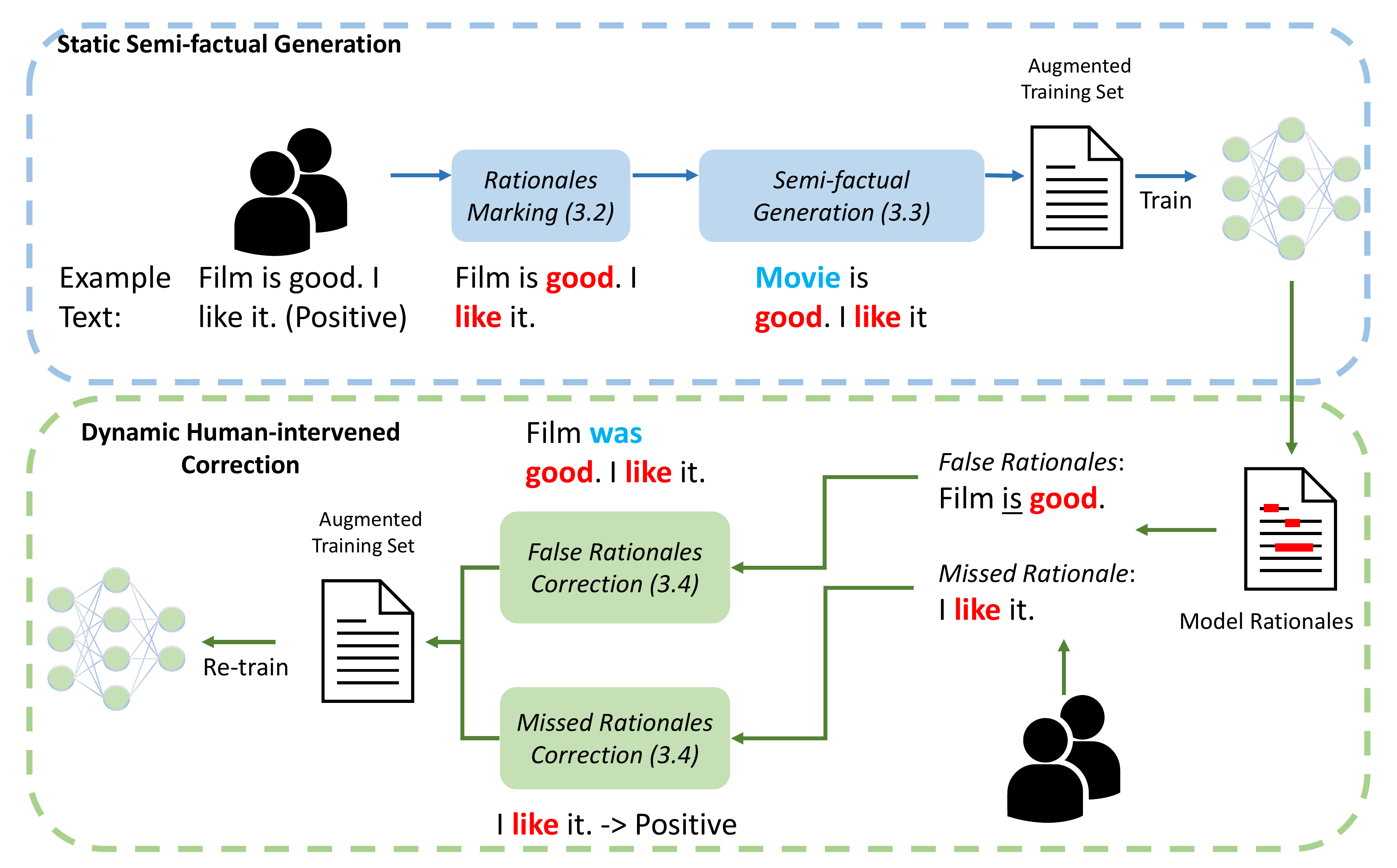}
 \caption{The procedure of the Rationale-centric Double-robustness Learning framework. Red text highlights rationales identified by human annotators. Blue text indicates words replaced in raw text. Underlined text shows spurious patterns identified by the model.}\label{fig:overview}
\end{figure}


Our main idea is a rationale-centric strategy for eliminating the effect of spurious patterns by leveraging  human knowledge as shown in Figure \ref{fig:overview}. 
Our double-robustness framework consists of two main modules. The first is a \emph{Static Semi-factual Generation module} that generates a set of semi-factual data automatically for a given instance by using human-identified rationales. Such labelling requires less human input compared to fully manual counterfactual generation (see Section \ref{subsec:rationale_marking}). In contrast with counterfactuals  \cite{roese1997counterfactual} that rely on what might have been different (i.e. the label would be changed if certain terms have been changed), semi-factuals \cite{mccloy2002semifactual,kenny2021generating}, as used in our work, aim to guide a model to identify terms less causally related to the label (i.e. even if certain terms had been changed, the label would be kept the same). Second, we apply a \emph{Dynamic Human-intervened Correction module}, where the most salient features are identified for model predictions over a set of training examples, and human workers intervene by checking the correctness of the rationale in case  first-round modifications introduce new artefacts. We evaluate the two modules in a few-shot setting, where a minimum number of training instances are labeled for maximum generalisation power, both for in-distribution and OOD predictions.\looseness=-1

Results on a sentiment analysis task, which is also used in \citet{kaushik2020learning}, demonstrate that the double-robust models can be less sensitive to spurious patterns. In particular, models trained with RDL with only 50 labelled examples achieve the same or even better results than fully-supervised training with a full training set of 1,707 examples, and improvements are especially significant for OOD tests. The predictive model trained with RDL using only 100 labelled examples outperforms models trained with manual \citep{kaushik2020learning} and automatic CAD \citep{yang-etal-2021-exploring} using the full augmented training set of 3,414 examples.

To the best of our knowledge, we are the first to exploit the efficacy of semi-factuals and human-intervention for improving the generalisation abilities of deep neural networks in few-shot learning scenarios.\footnote{All resources are available at https://github.com/GeorgeLuImmortal/RDL-Rationales-centric-Double-robustness-Learning/}



\section{Related Work}\label{sec:related}


\textbf{Data augmentation} has been used for resolving artefacts  in training datasets before \citep{gururangan2018annotation,srivastava2020robustness,kaushik2021learning}. In particular, previous work \cite{kaushik2020learning} relied on large-scale crowd-sourcing to generate useful augmented data. More recently, \citet{yang-etal-2021-exploring}, and \citet{wang2021robustness} investigated the efficacy of the automatically generated counterfactuals for sentiment analysis. Similar to our work, these methods also consider the most salient features that a model uses when generating augmented data, which is in line with our rationale definition. However, they use sentiment lexicon matching for identifying rationales, which is task-specific and not necessarily fully relevant. In contrast, we employ human annotators to identify rationales, which can be task-agnostic and robust. Moreover, our method generates semi-factuals instead of counterfactuals used in previous work. 


\noindent \textbf{Human-the-loop Machine Learning} \citep{wu2021survey} has received increasing research attention. Active learning \citep{settles2009active,margatina2021active}, the most common example of human-in-the-loop machine learning, asks human annotators only to provide high-level annotations (i.e. labels) for important examples.
There is also some work exploring more explainable AI systems by exploiting feature-based information. Such methods use relatively simple models such as Naïve Bayes \citep{Stumpf2009,kulesza2015} and Linear Regression with bag-of-words features \citep{JiaL17,teso2019,ghai2021,Shao_Skryagin_Stammer_Schramowski_Kersting_2021}, because these classifiers are relatively intuitive in generating explanations and amenable to incorporating human feedback. 

Some other work uses simple neural networks such as multi-layer perceptrons \citep{Shao_Skryagin_Stammer_Schramowski_Kersting_2021} and shallow CNNs \citep{lertvittayakumjorn2020find,StammerSK21,teso2021interactive} because the predictions of such models can be explained in the form of features. Very recently, \citet{yao2021refining} proposed a human-in-the-loop method to inspect more complicated models (e.g. BERT) with the help of model-agnostic post-hoc explanation algorithms \citep{ribeiro2018anchors} that can explain predictions of any linear or non-linear model without exploiting its weights. However, previous work focuses on increasing the explainability of AI systems for high-stakes domains such as health and finance \cite{li2020maec,yang2020html}, instead of improving model robustness or generalisation ability. Also, they assume access to a large amount of labelled data. In contrast, we focus on few-shot learning scenarios which are more compelling.

\section{Method}\label{sec:methods}

The RDL pipeline is shown in Figure \ref{fig:overview} and consists of two modules: \emph{Static Semi-factual Generation} and \emph{Dynamic Human-intervened Correction}. 

Static semi-factual generation is a more efficient alternative to manually generated counterfactuals \citep{kaushik2020learning}. In the first phase, Rationale Marking (Section \ref{subsec:rationale_marking}), human annotators review each document in the training set to provide \emph{rationales} (i.e. phrases that support the document classification decisions shown as bold text in Figure \ref{fig:overview}). The second phase is a semi-factual generation method based on synonym replacement (Section \ref{subsec:nsr_og}) that produces augmented examples (blue text in Figure \ref{fig:overview} indicates replaced words), which are added into the training set.

Dynamic human-intervened correction (Section \ref{subsec:human_correct}) is a rationales-powered human-in-the-loop framework to dynamically correct the model's behaviours. At the outset, \emph{sampling and sensitivity of contextual decomposition} (SCD) \citep{jin2019towards} is applied to detect the rationales given by the model that is obtained in the previous step. Then, all model-identified rationales (underlined texts in Figure \ref{fig:overview}) are examined by human annotators to identify \emph{false rationales} (i.e. words or phrases that do not support the classifications but are falsely included by the model) and \emph{missing rationales} (i.e. words or phrases that support the classifications but are not included by the model). Both false rationales and missing rationales are corrected to produce augmented examples. Finally, newly generated examples are added into the training set to re-train the deep learning model.

\subsection{Rationale Marking}\label{subsec:rationale_marking}


Following \citet{kaushik2020learning} and \citet{yang-etal-2021-exploring}, we use the \emph{IMDb} movie review dataset \citep{maas-etal-2011-learning} in our experiments. It consists of positive and negative movie reviews that are easy for human participants to understand, re-annotate, and provide feedback upon \cite{zaidan-etal-2007-using}. 

We use a crowdsourcing company to recruit editors and annotators for marking rationales that support classification decisions. At the outset, annotators were given instructions and examples that gently guided them to annotate rationales. Only adjectives, adverbs, nouns, and verbs were considered as rationales. Besides, rationales were required to carry complete semantic information. For example, for a phrase starting with a negation word such as ``\emph{not great}'', annotators are instructed to mark the whole phrase ``\emph{not great}'' as a rationale instead of just marking ``\emph{not}''. We also limited rationales to at most three consecutive words (i.e. unigrams, bigrams and trigrams). Phrases consisting of numerical scores are not counted as rationales (e.g. 5 or 10 stars) since different datasets may use different rating scales, and annotating digits may hurt OOD performance.\looseness=-1




Overall, we encouraged annotators to try their best to mark as many rationales as possible to explain classification labels. However, to guarantee the quality of rationale marking and prevent annotators from over including non-rationales for more payment, we also manually inspected annotated examples and rejected examples that contained incorrect rationales. After inspection, we rejected 10.6\% of negative reviews and 7.6\% of positive reviews. Editors and annotators re-annotated the rejected examples, which were then presented to us for another inspection. All re-annotated examples were approved only if all authors were happy with the quality of the annotations. Otherwise, the examples were re-annotated again.

Our annotation procedure generated 5,073 rationales in 855 movie reviews involved in Section \ref{subsec:rationale_marking} and \ref{subsec:human_correct}  (note that we did not annotate all 1,707 examples in the training set because only 855 examples were necessarily involved in our experiments). Human annotators spent on average 183.68 seconds to identify rationales in a review and our method generated semi-factual examples automatically. On the contrary, workers spent on average 300 seconds to revise a review to generate a counterfactual manually as reported by \citet{kaushik2020learning}. Note that our approach using 100 labelled examples can outperform manual CAD \citep{kaushik2020learning} using the entire training set of 1,707 examples (see Section \ref{subsec:exp_rp}), making our approach $\frac{300\times1707}{183.68\times100}\approx 27.88$ times more efficient than manually generated CAD. \looseness=-1

\subsection{Static Semi-factual Generation} \label{subsec:nsr_og}

We take a simple replacement strategy, which has been taken by \citet{yang-etal-2021-exploring}, to generate semi-factual examples. Given a human-identified rationale, our method constructs augmented examples by automatically replacing non-rationale words, thus leading to examples with the same labels. This augmentation is consistent with semi-factual thinking: even if those non-rationales were changed, the label would not change.

Formally, given a training example $x_{i} = [t_{i1}, t_{i2}, ... , t_{ij}]$ (where $t_{ij}$ is the $j^{th}$ token of the $i^{th}$ document) and its ground truth label $y_{i}$, we create a rationale vector $r_{i} = [a_{i1}, a_{i2}, ..., a_{ij}]$ where $a_{ij}$ is the value that indicates whether $t_{ij}$ is a rationale or not (we set $a_{ij}=1$ to indicate that $t_{ij}$ is a rationale and $0$ otherwise). To generate a semi-factual example, $x_{i}^{\prime}$, we randomly replace a certain number of non-rationales (where $a_{ij}=0$), except for punctuation, with synonymous terms. The synonyms can be provided by a human, retrieved automatically from a lexicon such as WordNet \citep{10.1145/219717.219748}, or generated using the \emph{mask-filling} function of a pretrained context-aware language model \citep{Liu2019RoBERTaAR}. \looseness=-1

In our experiments, we randomly replace 5\% of non-rationales using mask-filling and generate a set of augmented examples, $x_{i}^{\prime}$, with some replaced non-rationales and all the other tokens identical to $x_{i}$. The label, $y_{i}$, of a newly generated example is the same as the label of the original example, $x_{i}$. Examples of generated data are shown in Table \ref{tab:example_non_rationale_replacement}. Afterwards, the augmented examples are added into the training set used to train the model.

\begin{table*}[!t]
\centering
\small
\begin{tabular}{ll}
\hline
\textbf{Sentiment} & \textbf{Examples}                                          \\ \hline
Negative         & Origin: The attempt at a "lesbian scene" was \textbf{sad}. \\
                 & Augment 1: The \textcolor{blue}{hint} at a "lesbian scene" was \textbf{sad} .           \\
                 & Augment 2: The attempt at a "\textcolor{blue}{kiss} scene" was \textbf{sad} .           \\\hline
Positive         & Origin: I \textbf{recommended} this film a lot, specially in this difficult times for the planet . \\
                 & Augment 1: I \textbf{recommended} \textcolor{blue}{you} film a lot, specially in this difficult times for the planet .           \\
                 & Augment 2: I \textbf{recommended} this \textcolor{blue}{movie} a lot, specially in this difficult times for the planet .      \\
                 \hline
\end{tabular}
\caption{Fragments of augmented data generated by static semi-factual generation (Original/Augmented, in order). Blue spans were synonyms used as replacements and bold font were rationales identified by human annotators. }
\label{tab:example_non_rationale_replacement}
\end{table*}

\begin{table*}[!t]
\centering
\small
\begin{tabular}{ll}
\hline
Sentiment & Examples                                          \\ \hline
Negative         & Origin:  but this is \textbf{pathetic}! Micawber was nothing more than a mid-nineteenth century Kramer.  \\
                 & SCD: but this is \textbf{pathetic}! \underline{Micawber was} nothing more than a mid-nineteenth century Kramer.           \\
                 & Augment 1: but this is \textbf{pathetic}! \textcolor{blue}{Perkins became} nothing more than a mid-nineteenth century Kramer.           \\
                 & Augment 2: but this is \textbf{pathetic}! \textcolor{blue}{It had} nothing more than a mid-nineteenth century Kramer.           \\\hline
Positive         & Origin: Soylent Green is a wild movie that I \textbf{enjoyed} very much . \\
                 & SCD: \underline{Soylent Green} is a wild movie that I \textbf{enjoyed} very much .            \\
                 & Augment 1: \textcolor{blue}{Gang Orange} is a wild movie that I \textbf{enjoyed} very much .            \\
                 & Augment 2: \textcolor{blue}{Village Spring} is a wild movie that I \textbf{enjoyed} very much .       \\
                 \hline
\end{tabular}
\caption{Fragments of augmented data generated by false rationale correction (Original/SCD/Augmented, in order). Underlined spans were false rationales given by the model through SCD. Blue spans were synonyms used as replacements, and bold font were rationales identified by human annotators. }
\label{tab:example_false_rationale_replacement}
\end{table*}

\begin{table*}[t]
\centering
\small
\begin{tabular}{p{0.3\textwidth}p{0.1\textwidth}p{0.1\textwidth}p{0.1\textwidth}p{0.1\textwidth}p{0.1\textwidth}}
\hline
\textbf{Training Data} & \multicolumn{1}{l}{\textbf{In-domain}} & \multicolumn{1}{l}{\textbf{SemEval-2017}} & \multicolumn{1}{l}{\textbf{SST-2}} & \multicolumn{1}{l}{\textbf{Yelp}} & \textbf{Amazon} \\ \hline
Static (50 gold)     & 88.60{\scriptsize\emph{±1.11}}                          & 77.28{\scriptsize\emph{±9.11}}                  & 79.29{\scriptsize\emph{±5.14}}                & 91.53{\scriptsize\emph{±2.06}}               &   89.63{\scriptsize\emph{±1.65}}     \\
Full (1,707 gold)   &  \textbf{93.23{\scriptsize\emph{±0.46}}}          & 71.17{\scriptsize\emph{±2.54}}          & 80.23{\scriptsize\emph{±2.09}}          & \textbf{93.66{\scriptsize\emph{±0.84}}}          & 90.29{\scriptsize\emph{±0.57}} \\
\hline
DP (Static + 350 auto) (400) & 86.70{\scriptsize\emph{±2.92}}                          & 74.36{\scriptsize\emph{±2.92}}                  & 77.33{\scriptsize\emph{±6.01}}                & 89.60{\scriptsize\emph{±2.51}}               &   89.15{\scriptsize\emph{±1.89}}   \\
RR (Static + 350 auto) (400) & 89.65{\scriptsize\emph{±1.27}}           & 79.20{\scriptsize\emph{±1.27}}                  & 78.89{\scriptsize\emph{±5.95}}                & 91.93{\scriptsize\emph{±2.10}}               &  89.73{\scriptsize\emph{±1.26}}      \\
 \hline
\textbf{Our Methods}&&&&&
\\\hline
Static + 150 auto (200)    & 90.08{\scriptsize\emph{±1.25}}                          & 78.88{\scriptsize\emph{±6.67}}                  & 79.40{\scriptsize\emph{±3.28}}                & 92.19{\scriptsize\emph{±1.51}}               &    89.81{\scriptsize\emph{±1.73}}    \\ 
Static + 350 auto (400)    & 90.16{\scriptsize\emph{±0.85}}               & 80.54{\scriptsize\emph{±2.81}}                  & \textbf{81.26{\scriptsize\emph{±1.97}}}       & 93.03{\scriptsize\emph{±1.08}}      &  90.09{\scriptsize\emph{±1.79}}      \\ 
Static + 550 auto (600)    & 90.04{\scriptsize\emph{±1.50}}                         & 80.69{\scriptsize\emph{±3.42}}                  & 81.23{\scriptsize\emph{±1.83}}                & 92.10{\scriptsize\emph{±3.07}}               &    89.67{\scriptsize\emph{±1.27}}    \\ 
Static + 750 auto (800)     & 90.08{\scriptsize\emph{±1.01}}                          & 80.55{\scriptsize\emph{±3.96}}                  & 80.75{\scriptsize\emph{±2.30}}                & 92.36{\scriptsize\emph{±1.87}}               &   90.18±{\scriptsize\emph{1.44}}     \\ 
Static + 950 auto (1000) 
& 89.83{\scriptsize\emph{±1.28}}                          & \textbf{80.90{\scriptsize\emph{±3.29}}}         & 80.58{\scriptsize\emph{±2.57}}                & 92.30{\scriptsize\emph{±2.19}}               &    \textbf{90.62{\scriptsize\emph{±1.29}}}    \\ 
Static + 1150 auto (1200)         & 90.12{\scriptsize\emph{±1.82}}                          & 79.31{\scriptsize\emph{±1.82}}                  & 79.52{\scriptsize\emph{±3.15}}                & 91.47{\scriptsize\emph{±3.61}}               &    90.16±{\scriptsize\emph{1.46}}    \\ \hline
\end{tabular}
\caption{Results on in-distribution and OOD data. Values in brackets are the training set size. Static: uses 50 gold examples. Full: uses the full training set. Static + $n$: our static semi-factual generation method where $n$ is the number of semi-factuals. RR: Random Replacement \citep{wei-zou-2019-eda}. DP: Duplication.}
\label{tab:na_vs_orig}
\end{table*}

\subsection{Dynamic Human-intervened Correction}\label{subsec:human_correct}

Dynamic human-intervened correction further improves the robustness of the model by allowing human annotators to correct the model rationales online.
Firstly, SCD is applied to detect unigrams, bigrams or trigrams that are salient to the model. SCD is a technique to assess the importance of terms by continuously removing terms and measuring changes in prediction \citep{jin2019towards}. Human annotators examine all rationales given by the model from all documents to discover two types of incorrect rationale: false rationales and missing rationales.
The next phase allows human feedback to influence the learning process. To this end, for each type of incorrect rationale, we propose a corresponding strategy to correct them.

For false rationales (i.e. phrases that actually do not support classifications but are incorrectly identified by the model), we use synonym replacement again to generate semi-factual examples. Unlike the static semi-factual generation (Section \ref{subsec:nsr_og}), in this component we replace all false rationales with their synonyms instead of randomly replacing 5\% of non-rationales in a document. Examples of generated data are shown in Table \ref{tab:example_false_rationale_replacement}.

For missing rationales (i.e. phrases that actually support classifications but are not identified by the model), we take another simple semi-factual generation strategy, that is, extracting sentences that contain missing rationales to form semi-factual data. Specifically, given a sentence containing missing rationales, we use this sentence as a new example, and the label of this newly generated example is identical to that of the document where the sentence is extracted. For example, there is a positive movie review (bold font for rationales) \emph{``Robert Urich was a \textbf{fine} actor, and he makes this TV movie \textbf{believable} . I remember watching this film when I was 15 ....''}. The model fails to identify \emph{``\textbf{fine}''} and \emph{``\textbf{believable}''} as rationales. Thus we extract the text \emph{````Robert Urich was a \textbf{fine} actor, and he makes this TV movie \textbf{believable} .''} as a new example, and the class of this example is still positive. We extract the whole sentence rather than just the missing rationales to reserve more semantic information. 

Note that the two correction methods in dynamic human-intervened correction can operate in parallel and the generated examples are added to the small training set to re-train the model.


\section{Why Does RDL Work?}


Broadly speaking, our RDL framework takes advantage of invariance that makes a model less sensitive to non-rationale words or spurious patterns \citep{tu2020empirical,wang2021identifying} in favour of focusing on useful mappings of rationales to labels. 


More specifically, by using static semi-factual generation (Section \ref{subsec:nsr_og}) and false rationale correction (Section \ref{subsec:human_correct}), we expect to break spurious associations. For example, if a model incorrectly determines that ``\emph{Soylent Green}'' is associated with positive sentiment (Table \ref{tab:example_false_rationale_replacement}), the augmented examples that replace ``\emph{Soylent Green}'' with other phrases such as ``\emph{Gang Orange}'' break the spurious association. Besides, using synonym replacement can generate examples that are similar to the original one, which is equivalent to adding noisy data to prevent models from overfitting \citep{wei-zou-2019-eda}. \looseness=-1

Missing rationale correction (Section \ref{subsec:human_correct}) emphasizes the ground truth associations between rationales and labels, enabling the model to better estimate the generally useful underlying distributions for OOD datasets, even in few-shot learning scenarios. In the next section, we present experiments and empirical evidence to demonstrate the utility of the proposed RDL framework in improving model robustness.

\section{Experiments}

Our intention is to improve the generalisability of models, and we use both in-distribution and OOD performance for evaluation. Our experiments are designed to address the following research questions: 
\begin{itemize}
\item \textbf{RQ1} Can we use static semi-factual generation to achieve better in-distribution and OOD performance? 
\item \textbf{RQ2} Does dynamic human-intervened correction improve generalisability of models?
\end{itemize}
\looseness=-1

\subsection{Datasets}

For fair comparison with previous work \citep{kaushik2020learning,yang-etal-2021-exploring}, we  use the \emph{IMDb} sentiment classification dataset \citep{maas-etal-2011-learning} as the in-distribution dataset. Following \citet{kaushik2020learning}, all models were trained with the \emph{IMDb} dataset predefined training, validation and test partitions containing $1,707$, $245$, and $488$ reviews respectively and an enforced 50:50 class ratio.



To measure the generalisation ability of different models, we focus on OOD performance. To this end, we test models on another four binary sentiment classification datasets: the sampled \emph{Amazon reviews} dataset \citep{ni-etal-2019-justifying} (100,000 positives and 100,000 negatives) from six genres: beauty, fashion, appliances, gift cards, magazines, and software; the \emph{Yelp review} dataset \citep{zhang2015character} (19,000 positives and 19,000 negatives); the \emph{SST-2} dataset \citep{socher-etal-2013-recursive} (1,067 positives and 1,143 negatives), and the \emph{SemEval-2017 Twitter} dataset \citep{rosenthal-etal-2017-semeval} (2,339 positives and 2,339 negatives). These datasets were sampled to ensure a nearly 50:50 class balance.

\subsection{Evaluating Static Semi-factual Generation}\label{subsec: experiment_1}

To address \textbf{RQ1}, we compare the performance of models trained by the \textbf{static semi-factual generation} strategy with models trained with the original 50 examples, referred to as \textbf{Static}. We also compare to a model trained with the full training set (1,707 labelled examples), referred to as \textbf{Full}.

\subsubsection{Experiment Setup} \label{subsec: experiment_1_setup}

To simulate the few-shot training scenario, we randomly sample 50 examples (we also forced a 50:50 class balance) from the \emph{IMDb} dataset as training data. For each experiment, the training is repeated 10 times with training datasets sampled by 10 different random seeds. We report the average result of these 10 repetitions and use accuracy to measure the classification performance. Our experiments rely on an off-the-shelf cased ``RoBERTa-base'' model implemented by Hugging Face\footnote{https://huggingface.co/transformers/model\_doc/roberta.html} to either perform mask-filling to provide synonyms or as a predictive model. Following \citet{kaushik2020learning}, we fine-tune RoBERTa for up to 20 epochs and apply early stopping with patience of 5 (i.e. stop fine-tuning when validation loss does not decrease for 5 epochs). \looseness=-1

We also explore the impact of the number of semi-factual examples on model performance. To this end, we conduct static semi-factual generation with a different number of augmented examples for each instance: \{3, 7, 11, 15, 19, 23\}. Considering we have 50 original examples, this would result in \{150, 350, 550, 750, 950, 1,150\} additional examples in the training set, respectively (we call this \textbf{Static+\emph{n}}, where \emph{n} is the number of generated semi-factuals). \looseness=-1

We use the Adam optimizer \citep{adam} with a batch size of 4. We found that setting the learning rate to \{5e-5, 5e-6 and 5e-6\} could optimise Static, Static+\emph{n}, and Full, respectively.

\subsubsection{Results and Analysis}

As shown in Table \ref{tab:na_vs_orig}, all static semi-factual generation (Static+\emph{n}) methods can outperform the baseline method (Static) in both in-distribution and OOD tests, demonstrating the utility of static semi-factual generation. Among all Static+\emph{n} methods, Static+350 seems the best-performing method and exceeds Static with a 1.56\% in-distribution improvement in average accuracy. Static+350 also outperforms Static with 3.26\%, 1.97\%, 1.5\%, and 0.46\% OOD improvement in the \emph{SemEval-2017}, \emph{SST-2}, \emph{Yelp} and \emph{Amazon} datasets respectively. Although the improvement on the \emph{Amazon} dataset appears modest, given that there are 200,000 examples in the \emph{Amazon} test set, this actually stands for nearly 1,000 documents being correctly classified.

The Static+\emph{n} methods can even outperform Full (i.e. normal training with the full training set) on the \emph{SemEval}, \emph{SST-2}, and \emph{Amazon} datasets and are comparable on the \emph{Yelp} dataset. The performance of models with the full training set is best on the in-distribution dataset but the worst on the \emph{SemEval} dataset, which can be caused by the big difference between underlying distributions of these two datasets. In other words, a model that fits well with one dataset can cause performance decay on others. In this case, training with a smaller training set is more likely to reduce overfitting with the in-distribution dataset and fit well with the \emph{SemEval} dataset, which explains the big improvement. It is interesting to note that models trained with the entire training set perform slightly better on the OOD \emph{Yelp} dataset (93.66{\scriptsize\emph{±0.84}}) than on the in-distribution dataset (93.23{\scriptsize\emph{±0.46}}), which could also be explained by the high similarity between the underlying distributions of these two datasets.

\noindent \textbf{Benefits of Static Semi-factual Generation}

\noindent First, we test whether the improvement in model performance is brought about by static semi-factual generation (Static+\emph{n}) or simply by an increase in the size of the training set. We compare Static+350 (due to its relatively good performance) with another baseline called Duplication (\textbf{DP} heareafter). We multiply the original training set (50 examples) up into 400 examples identical to the size of the training set of Static+350, and fine-tune RoBERTa on this dataset with the same hyperparameters as Static+350.\looseness=-1

As shown in Table \ref{tab:na_vs_orig}, in most cases, DP underperforms other algorithms and is even worse than Static, demonstrating that solely increasing the dataset size cannot improve the performance. We believe that the duplication of original examples increases the risk of overfitting and easily \emph{magnifies} artefacts or spurious patterns hidden in the small training set, which leads to worse models.

Second, synonym replacement has been used previously for data augmentation \citep{wei-zou-2019-eda}, and we compare static semi-factual generation with simply replacing any words (i.e. both rationales and non-rationales). Following \citet{wei-zou-2019-eda}, we replace 5\% of words at random and set the training set size to 400 to ensure fair comparison (we use RoBERTa and the same hyperparameters of Static+350). We call this Random Replacement (\textbf{RR} hereafter). \looseness=-1


As shown in Table \ref{tab:na_vs_orig}, RR is slightly better than the baseline Static approach. This result is similar to that reported in \citet{wei-zou-2019-eda}, since the augmented data generated by random replacement is similar to the original data, introducing noise that helps prevent overfitting to some extent. However, the magnitude of improvement of the Static+\emph{n} method is much larger than that of RR, demonstrating the utility of only replacing non-rationales to generate semi-factuals. These observations show that the model trained with Static+\emph{n} does improve both in-distribution and OOD performance, and the improvement is actually derived from static semi-factual generation.

\begin{figure}[!t]
\centering
\includegraphics[width=.35\textwidth]{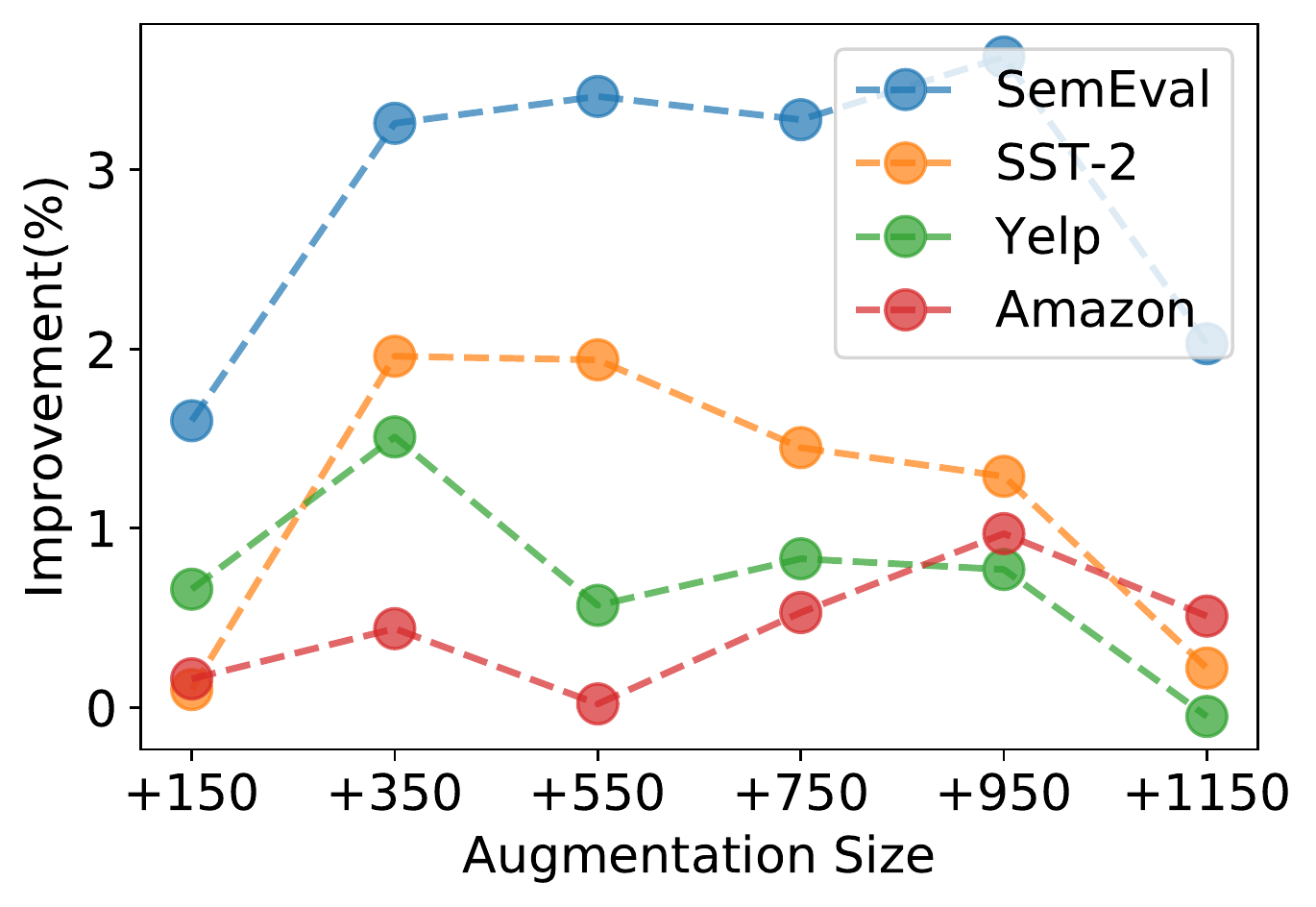}
 \caption{Average performance gain of different static semi-factual generation methods with different augmentation size over four OOD datasets.}\label{fig:improvement}
\end{figure}

\begin{table*}[!t]
\centering
\small
\begin{tabular}{p{0.33\textwidth}p{0.1\textwidth}p{0.1\textwidth}p{0.1\textwidth}p{0.1\textwidth}p{0.1\textwidth}}
\hline
\textbf{Baseline Methods} & \multicolumn{1}{l}{\textbf{In-domain}} & \multicolumn{1}{l}{\textbf{SemEval-2017}} & \multicolumn{1}{l}{\textbf{SST-2}} & \multicolumn{1}{l}{\textbf{Yelp}} & \textbf{Amazon} \\ \hline
Static (50 gold)     & 88.60{\scriptsize\emph{±1.11}}                          & 77.28{\scriptsize\emph{±9.11}}                  & 79.29{\scriptsize\emph{±5.14}}                & 91.53{\scriptsize\emph{±2.06}}               &    89.63{\scriptsize\emph{±1.65}}    \\
Static + 350 auto (400)    & 90.16{\scriptsize\emph{±0.85}}                 & 80.54{\scriptsize\emph{±2.81}}                  & 81.26{\scriptsize\emph{±1.97}}       & 93.03{\scriptsize\emph{±1.08}}      &    90.09{\scriptsize\emph{±1.79}}    \\
AL (100 gold)      & 88.64{\scriptsize\emph{±1.75}}                          & 78.61{\scriptsize\emph{±5.90}}                  & 80.50{\scriptsize\emph{±3.37}}                & 92.47{\scriptsize\emph{±0.68}}               &    89.80{\scriptsize\emph{±1.91}}    \\\hline
\textbf{CAD-based Methods}      &                           &          &                 &                &  \\\hline
Manual CAD (3,414 gold)      & \textbf{92.70{\scriptsize\emph{±0.53}}}          & 69.98{\scriptsize\emph{±3.99}}          & 80.30{\scriptsize\emph{±2.03}}        & 91.87{\scriptsize\emph{±1.09}}&90.48{\scriptsize\emph{±1.09}}  \\

Automatics CAD (1,707 gold+1,707 auto) & 91.82{\scriptsize\emph{±0.74}} & 79.39{\scriptsize\emph{±5.37}}     &  80.60{\scriptsize\emph{±3.10}}    &  91.92{\scriptsize\emph{±0.97}}  & 90.46{\scriptsize\emph{±1.08}} \\\hline 
\textbf{Our Dynamic Methods}      &                           &          &                 &                &  \\
\hline

Dynamic (100 gold + 700 auto)  & 90.84{\scriptsize\emph{±0.99}}                          & 80.32{\scriptsize\emph{±4.31}}                  & \textbf{82.40{\scriptsize\emph{±2.14}}}       & \textbf{93.19{\scriptsize\emph{±1.24}}}      &       \textbf{90.51{\scriptsize\emph{±2.17}}} \\
Dynamic-MR (100 gold + 700 auto)        & 91.06{\scriptsize\emph{±1.21}}                 & 79.04{\scriptsize\emph{±4.92}}                  & 82.24{\scriptsize\emph{±2.59}}                & 93.03{\scriptsize\emph{±1.92}}               & 
90.22{\scriptsize\emph{±2.74}}\\
Dynamic-FR (100 gold + 700 auto)        & 89.85{\scriptsize\emph{±1.38}}                          & \textbf{82.39{\scriptsize\emph{±1.88}}}         & 81.59{\scriptsize\emph{±1.82}}                & 92.98{\scriptsize\emph{±0.91}}               & 90.12{\scriptsize\emph{±2.42}} \\ \hline

\end{tabular}
\caption{Results on in-distribution and OOD data. Values in brackets are the training set size. AL: Active Learning. Manual CAD \citep{kaushik2020learning}, Automatic CAD \citep{yang-etal-2021-exploring}. Our methods are Dynamic-MR: Missing Rationale Correction, Dynamic-FR: False Rationale Correction, Dynamic: Dynamic Human-intervened Correction. }
\label{tab:hybrid_og}
\end{table*}

\subsection{Evaluating Dynamic Human-intervened Correction}\label{subsec:exp_rp}

As shown in Table \ref{tab:na_vs_orig} and Figure \ref{fig:improvement}, the performance gain of static semi-factual generation (Static+\emph{n}) marginalises when augmented data is increased. Using too much augmented data even hurts the Static+1150 performance. This observation is consistent with existing work on data augmentation \citep{wei-zou-2019-eda}. We believe one reason could be that the use of static augmented examples could also introduce new spurious patterns that degrade model performance, necessitating a method that exploits rationales without generating too many augmented examples. Human-in-the-loop can address this issue by dynamically correcting the model.

To address \textbf{RQ2}, we compare the performance of models trained by \textbf{dynamic human-intervened correction} with a popular few-shot human-in-the-loop learning framework, Active Learning, as well as two other state-of-the-art CAD-based methods \citep{kaushik2020learning,yang-etal-2021-exploring}. Lastly, we provide an ablation study to examine the influence of different correction methods, as well as an analysis regarding model sensitivity to spurious patterns. \looseness=-1

\subsubsection{Experiment Setup}  

We build up an active learning procedure as a baseline based on the model trained with Static. In particular, we select another 50 examples by Uncertainty Sampling (i.e. prediction scores for two classes in these examples were close) and add them into the training set (called \textbf{AL} hereafter). The training set size of the baseline becomes 100. The best performing static semi-factual generation method Static+350 is also listed as a baseline.

For fair comparison, we also use Uncertainty Sampling to select another 50 examples (i.e. 100 original examples in the training set now) for the proposed dynamic human-intervened correction including both False Rationale Correction and Missing Rationale Correction (called \textbf{Dynamic}). For Dynamic, we control the number of augmented examples for each review to 7 (4 from Missing Rationale Correction and 3 from False Rationale Correction), resulting in 800 examples in the training set. For Automatic CAD \citep{yang-etal-2021-exploring} and Manual CAD \citep{kaushik2020learning}, we use the entire training set to produce counterfactuals to build up two challenging baselines (one counterfactual for one example, which is limited by the method), resulting in 3,414 examples in the training set.

To investigate the influence of each correction method, we also construct another two datasets that augment the same 100 original examples to 800 exclusively by False Rationale Correction (\textbf{Dynamic-FR} hereafter) and Missing Rationale Correction (\textbf{Dynamic-MR} hereafter). Again, experiments all rely on a RoBERTa model and all hyperparameters are identical to those described in Section \ref{subsec: experiment_1_setup}, except for the learning rate of AL which is set to 1.25e-5 (we found this value optimised AL performance).

\subsubsection{Results and Analysis}

As shown in Table \ref{tab:hybrid_og}, 
both AL and Dynamic outperform Static in in-distribution and OOD datasets which makes sense, because we use Uncertainty Sampling to add new labelled data to minimise model uncertainty and increase model performance. However, 
AL fails to compete with Static+350 even if more original data is added, which again demonstrates the utility of static semi-factual generation. On the contrary, Dynamic does better than Static+350 with a 0.68\% in-distribution improvement in average accuracy. Dynamic also outperforms Static+350 with 1.14\%, 0.16\%, 0.42\% OOD improvement in the \emph{SST-2}, \emph{Yelp} and \emph{Amazon} datasets, but no improvement for the \emph{SemEval} dataset. Finally, the performance of our methods is better that the state-of-the-art manual CAD method in few-shot learning scenarios on all OOD datasets.

Overall, these observations demonstrate that applying dynamic human-intervened correction (i.e. Missing Rationale Correction and False Rationale Correction) can further increase the robustness of a model on generalisation ability, effectively avoiding the improvement marginalisation caused by the increased volume of augmented data.   

\noindent \textbf{Missing Rationales vs. False Rationales}

\noindent We conduct an ablation study by examining the performance of Dynamic-MR and Dynamic-FR in Table \ref{tab:hybrid_og}. Interestingly, Dynamic-FR is specifically good at improving model performance on the in-distribution and \emph{SemEval} datasets while Dynamic-MR does a good job on the \emph{SST-2} dataset. We believe that it is because Dynamic-MR biases the model to estimate an underlying distribution that is useful for \emph{SST-2} and in-distribution datasets, while Dynamic-FR biases the model to estimate a distribution similar to \emph{SemEval} dataset. The performance of Dynamic can be explained as a compromise of two correction methods.\looseness=-1

\noindent \textbf{Sensitivity to Spurious Patterns}

\begin{table}[t]
\centering
\small
\begin{tabular}{lll}
\hline
         & \textbf{Non-rationales} & \textbf{Rationales}     \\ \hline
Static & 0.572          & 0.428          \\
Dynamic & 0.433  & 0.567 \\ \hline
\end{tabular}
\caption{Static versus Dynamic models on average sensitivity (normalised) to rationales and non-rationales for \emph{IMDb} test samples.
}\label{tab:sensitivity}
\end{table}

\noindent We conduct an analysis to explore whether the double-robust models are less sensitive to spurious patterns. We compute models mean sensitivity to all rationales and non-rationales through SCD in the \emph{IMDb} test set. As shown in Table \ref{tab:sensitivity}, the corrected model is much more sensitive to rationales with 13.9\% average increase in the sensitivity to rationales, which demonstrates that our double-robust method can decouple models from spurious patterns. \looseness=-1

\section{Conclusion}

We proposed a rationale-centric human-in-the-loop framework, RDL, for better model generalisability in few-shot learning scenarios. Experimental results show that our method can boost performance of deep neural networks in both in-distribution and OOD datasets and make models less sensitive to spurious patterns, enabling fast generalisation. In the future, we expect to see rationale-centric frameworks defined for  different tasks, including NER, question answering, and relation extraction.\looseness=-1

\section{Ethical Statement}
We honor the ACL Code of Ethics. No private data or non-public information was used in this work. All annotators have received labor fees corresponding to the amount of their annotated instances. 

\section*{Acknowledgements}
We acknowledge with thanks the discussion with Chenyang Lyu from Dublin City University, as well as the many others who have helped. We would also like to thank anonymous reviewers for their insightful comments and suggestions to help improve the paper. This publication has emanated from research conducted with the financial support of the Pioneer and "Leading Goose" R\&D Program of Zhejiang under Grant Number 2022SDXHDX0003 and Science Foundation Ireland (SFI) under Grant Number [12/RC/2289\_P2]. Yue Zhang is the corresponding author.


\bibliography{anthology,custom}

\begin{thebibliography}{47}
\expandafter\ifx\csname natexlab\endcsname\relax\def\natexlab#1{#1}\fi

\bibitem[{Delaney et~al.(2021)Delaney, Greene, and
  Keane}]{delaney2021uncertainty}
Eoin Delaney, Derek Greene, and Mark~T Keane. 2021.
\newblock Uncertainty estimation and out-of-distribution detection for
  counterfactual explanations: Pitfalls and solutions.
\newblock \emph{arXiv preprint arXiv:2107.09734}.

\bibitem[{Feng et~al.(2021)Feng, Zhang, He, Zhang, and
  Chua}]{feng-etal-2021-empowering}
Fuli Feng, Jizhi Zhang, Xiangnan He, Hanwang Zhang, and Tat-Seng Chua. 2021.
\newblock \href {https://doi.org/10.18653/v1/2021.findings-acl.196} {Empowering
  language understanding with counterfactual reasoning}.
\newblock In \emph{Findings of the Association for Computational Linguistics:
  ACL-IJCNLP 2021}, pages 2226--2236, Online. Association for Computational
  Linguistics.

\bibitem[{Ghai et~al.(2021)Ghai, Liao, Zhang, Bellamy, and Mueller}]{ghai2021}
Bhavya Ghai, Q.~Vera Liao, Yunfeng Zhang, Rachel Bellamy, and Klaus Mueller.
  2021.
\newblock \href {https://doi.org/10.1145/3432934} {Explainable active learning
  (xal): Toward ai explanations as interfaces for machine teachers}.
\newblock \emph{Proc. ACM Hum.-Comput. Interact.}, 4(CSCW3).

\bibitem[{Gururangan et~al.(2018)Gururangan, Swayamdipta, Levy, Schwartz,
  Bowman, and Smith}]{gururangan2018annotation}
Suchin Gururangan, Swabha Swayamdipta, Omer Levy, Roy Schwartz, Samuel~R
  Bowman, and Noah~A Smith. 2018.
\newblock Annotation artifacts in natural language inference data.
\newblock \emph{arXiv preprint arXiv:1803.02324}.

\bibitem[{Jia and Liang(2017)}]{JiaL17}
Robin Jia and Percy Liang. 2017.
\newblock \href {https://doi.org/10.18653/v1/d17-1215} {Adversarial examples
  for evaluating reading comprehension systems}.
\newblock In \emph{Proceedings of the 2017 Conference on Empirical Methods in
  Natural Language Processing, {EMNLP} 2017, Copenhagen, Denmark, September
  9-11, 2017}, pages 2021--2031. Association for Computational Linguistics.

\bibitem[{Jin et~al.(2019)Jin, Wei, Du, Xue, and Ren}]{jin2019towards}
Xisen Jin, Zhongyu Wei, Junyi Du, Xiangyang Xue, and Xiang Ren. 2019.
\newblock Towards hierarchical importance attribution: Explaining compositional
  semantics for neural sequence models.
\newblock In \emph{International Conference on Learning Representations}.

\bibitem[{Kaushik et~al.(2020)Kaushik, Hovy, and Lipton}]{kaushik2020learning}
Divyansh Kaushik, Eduard Hovy, and Zachary~C Lipton. 2020.
\newblock Learning the difference that makes a difference with counterfactually
  augmented data.
\newblock \emph{International Conference on Learning Representations (ICLR)}.

\bibitem[{Kaushik et~al.(2021)Kaushik, Setlur, Hovy, and
  Lipton}]{kaushik2021learning}
Divyansh Kaushik, Amrith Setlur, Eduard Hovy, and Zachary~C Lipton. 2021.
\newblock Explaining the efficacy of counterfactually augmented data.
\newblock \emph{International Conference on Learning Representations (ICLR)}.

\bibitem[{Keith et~al.(2020)Keith, Jensen, and O’Connor}]{keith2020text}
Katherine Keith, David Jensen, and Brendan O’Connor. 2020.
\newblock Text and causal inference: A review of using text to remove
  confounding from causal estimates.
\newblock In \emph{Proceedings of the 58th Annual Meeting of the Association
  for Computational Linguistics}, pages 5332--5344.

\bibitem[{Kenny and Keane(2021)}]{kenny2021generating}
Eoin~M Kenny and Mark~T Keane. 2021.
\newblock On generating plausible counterfactual and semi-factual explanations
  for deep learning.

\bibitem[{Kingma and Ba(2014)}]{adam}
Diederik Kingma and Jimmy Ba. 2014.
\newblock Adam: A method for stochastic optimization.
\newblock \emph{International Conference on Learning Representations}.

\bibitem[{Kulesza et~al.(2015)Kulesza, Burnett, Wong, and Stumpf}]{kulesza2015}
Todd Kulesza, Margaret Burnett, Weng-Keen Wong, and Simone Stumpf. 2015.
\newblock \href {https://doi.org/10.1145/2678025.2701399} {Principles of
  explanatory debugging to personalize interactive machine learning}.
\newblock In \emph{Proceedings of the 20th International Conference on
  Intelligent User Interfaces}, IUI '15, page 126–137, New York, NY, USA.
  Association for Computing Machinery.

\bibitem[{Kulesza et~al.(2010)Kulesza, Stumpf, Burnett, Wong, Riche, Moore,
  Oberst, Shinsel, and McIntosh}]{kulesza2010}
Todd Kulesza, Simone Stumpf, Margaret Burnett, Weng-Keen Wong, Yann Riche,
  Travis Moore, Ian Oberst, Amber Shinsel, and Kevin McIntosh. 2010.
\newblock \href {https://doi.org/10.1109/VLHCC.2010.15} {Explanatory debugging:
  Supporting end-user debugging of machine-learned programs}.
\newblock In \emph{2010 IEEE Symposium on Visual Languages and Human-Centric
  Computing}, pages 41--48.

\bibitem[{Lertvittayakumjorn et~al.(2020)Lertvittayakumjorn, Specia, and
  Toni}]{lertvittayakumjorn2020find}
Piyawat Lertvittayakumjorn, Lucia Specia, and Francesca Toni. 2020.
\newblock \href {http://arxiv.org/abs/2010.04987} {Find: Human-in-the-loop
  debugging deep text classifiers}.

\bibitem[{Lertvittayakumjorn and
  Toni(2021)}]{lertvittayakumjorn2021explanation}
Piyawat Lertvittayakumjorn and Francesca Toni. 2021.
\newblock Explanation-based human debugging of nlp models: A survey.
\newblock \emph{arXiv preprint arXiv:2104.15135}.

\bibitem[{Li et~al.(2020)Li, Yang, Smyth, and Dong}]{li2020maec}
Jiazheng Li, Linyi Yang, Barry Smyth, and Ruihai Dong. 2020.
\newblock Maec: A multimodal aligned earnings conference call dataset for
  financial risk prediction.
\newblock In \emph{Proceedings of the 29th ACM International Conference on
  Information \& Knowledge Management}, pages 3063--3070.

\bibitem[{Liu et~al.(2019)Liu, Ott, Goyal, Du, Joshi, Chen, Levy, Lewis,
  Zettlemoyer, and Stoyanov}]{Liu2019RoBERTaAR}
Yinhan Liu, Myle Ott, Naman Goyal, Jingfei Du, Mandar Joshi, Danqi Chen, Omer
  Levy, Mike Lewis, Luke Zettlemoyer, and Veselin Stoyanov. 2019.
\newblock Roberta: A robustly optimized bert pretraining approach.
\newblock \emph{ArXiv}, abs/1907.11692.

\bibitem[{Lu et~al.(2021)Lu, Henchion, Bacher, and
  Namee}]{10.1007/978-3-030-88942-5_18}
Jinghui Lu, Maeve Henchion, Ivan Bacher, and Brian~Mac Namee. 2021.
\newblock A sentence-level hierarchical bert model for document classification
  with limited labelled data.
\newblock In \emph{Discovery Science}, pages 231--241, Cham. Springer
  International Publishing.

\bibitem[{Lu and MacNamee(2020)}]{lu2020investigating}
Jinghui Lu and Brian MacNamee. 2020.
\newblock Investigating the effectiveness of representations based on
  pretrained transformer-based language models in active learning for labelling
  text datasets.
\newblock \emph{arXiv preprint arXiv:2004.13138}.

\bibitem[{Maas et~al.(2011)Maas, Daly, Pham, Huang, Ng, and
  Potts}]{maas-etal-2011-learning}
Andrew~L. Maas, Raymond~E. Daly, Peter~T. Pham, Dan Huang, Andrew~Y. Ng, and
  Christopher Potts. 2011.
\newblock \href {https://aclanthology.org/P11-1015} {Learning word vectors for
  sentiment analysis}.
\newblock In \emph{Proceedings of the 49th Annual Meeting of the Association
  for Computational Linguistics: Human Language Technologies}, pages 142--150,
  Portland, Oregon, USA. Association for Computational Linguistics.

\bibitem[{Margatina et~al.(2021)Margatina, Vernikos, Barrault, and
  Aletras}]{margatina2021active}
Katerina Margatina, Giorgos Vernikos, Lo{\"\i}c Barrault, and Nikolaos Aletras.
  2021.
\newblock Active learning by acquiring contrastive examples.
\newblock \emph{Proceddings of the 2021 Conference on Empirical Methods in
  Natural Language Processing, Underline Science Inc.}

\bibitem[{McCloy and Byrne(2002)}]{mccloy2002semifactual}
Rachel McCloy and Ruth~MJ Byrne. 2002.
\newblock Semifactual “even if” thinking.
\newblock \emph{Thinking \& Reasoning}, 8(1):41--67.

\bibitem[{Miller(1995)}]{10.1145/219717.219748}
George~A. Miller. 1995.
\newblock \href {https://doi.org/10.1145/219717.219748} {Wordnet: A lexical
  database for english}.
\newblock \emph{Commun. ACM}, 38(11):39–41.

\bibitem[{Ni et~al.(2019)Ni, Li, and McAuley}]{ni-etal-2019-justifying}
Jianmo Ni, Jiacheng Li, and Julian McAuley. 2019.
\newblock \href {https://doi.org/10.18653/v1/D19-1018} {Justifying
  recommendations using distantly-labeled reviews and fine-grained aspects}.
\newblock In \emph{Proceedings of the 2019 Conference on Empirical Methods in
  Natural Language Processing and the 9th International Joint Conference on
  Natural Language Processing (EMNLP-IJCNLP)}, pages 188--197, Hong Kong,
  China. Association for Computational Linguistics.

\bibitem[{Qian et~al.(2021)Qian, Feng, Wen, Ma, and
  Xie}]{qian2021counterfactual}
Chen Qian, Fuli Feng, Lijie Wen, Chunping Ma, and Pengjun Xie. 2021.
\newblock Counterfactual inference for text classification debiasing.
\newblock In \emph{Proceedings of the 59th Annual Meeting of the Association
  for Computational Linguistics and the 11th International Joint Conference on
  Natural Language Processing (Volume 1: Long Papers)}, pages 5434--5445.

\bibitem[{Ribeiro et~al.(2018)Ribeiro, Singh, and
  Guestrin}]{ribeiro2018anchors}
Marco~Tulio Ribeiro, Sameer Singh, and Carlos Guestrin. 2018.
\newblock Anchors: High-precision model-agnostic explanations.
\newblock In \emph{Proceedings of the AAAI conference on artificial
  intelligence}, volume~32.

\bibitem[{Roese(1997)}]{roese1997counterfactual}
Neal~J Roese. 1997.
\newblock Counterfactual thinking.
\newblock \emph{Psychological bulletin}, 121(1):133.

\bibitem[{Rosenthal et~al.(2017)Rosenthal, Farra, and
  Nakov}]{rosenthal-etal-2017-semeval}
Sara Rosenthal, Noura Farra, and Preslav Nakov. 2017.
\newblock \href {https://doi.org/10.18653/v1/S17-2088} {{S}em{E}val-2017 task
  4: Sentiment analysis in {T}witter}.
\newblock In \emph{Proceedings of the 11th International Workshop on Semantic
  Evaluation ({S}em{E}val-2017)}, pages 502--518, Vancouver, Canada.
  Association for Computational Linguistics.

\bibitem[{Settles(2009)}]{settles2009active}
Burr Settles. 2009.
\newblock Active learning literature survey.

\bibitem[{Shao et~al.(2021)Shao, Skryagin, Stammer, Schramowski, and
  Kersting}]{Shao_Skryagin_Stammer_Schramowski_Kersting_2021}
Xiaoting Shao, Arseny Skryagin, Wolfgang Stammer, Patrick Schramowski, and
  Kristian Kersting. 2021.
\newblock \href {https://ojs.aaai.org/index.php/AAAI/article/view/17148} {Right
  for better reasons: Training differentiable models by constraining their
  influence functions}.
\newblock \emph{Proceedings of the AAAI Conference on Artificial Intelligence},
  35(11):9533--9540.

\bibitem[{Socher et~al.(2013)Socher, Perelygin, Wu, Chuang, Manning, Ng, and
  Potts}]{socher-etal-2013-recursive}
Richard Socher, Alex Perelygin, Jean Wu, Jason Chuang, Christopher~D. Manning,
  Andrew Ng, and Christopher Potts. 2013.
\newblock \href {https://www.aclweb.org/anthology/D13-1170} {Recursive deep
  models for semantic compositionality over a sentiment treebank}.
\newblock In \emph{Proceedings of the 2013 Conference on Empirical Methods in
  Natural Language Processing}, pages 1631--1642, Seattle, Washington, USA.
  Association for Computational Linguistics.

\bibitem[{Srivastava et~al.(2020)Srivastava, Hashimoto, and
  Liang}]{srivastava2020robustness}
Megha Srivastava, Tatsunori Hashimoto, and Percy Liang. 2020.
\newblock Robustness to spurious correlations via human annotations.
\newblock In \emph{International Conference on Machine Learning}, pages
  9109--9119. PMLR.

\bibitem[{Stammer et~al.(2021)Stammer, Schramowski, and Kersting}]{StammerSK21}
Wolfgang Stammer, Patrick Schramowski, and Kristian Kersting. 2021.
\newblock \href
  {https://openaccess.thecvf.com/content/CVPR2021/html/Stammer\_Right\_for\_the\_Right\_Concept\_Revising\_Neuro-Symbolic\_Concepts\_by\_Interacting\_CVPR\_2021\_paper.html}
  {Right for the right concept: Revising neuro-symbolic concepts by interacting
  with their explanations}.
\newblock In \emph{{IEEE} Conference on Computer Vision and Pattern
  Recognition, {CVPR} 2021, virtual, June 19-25, 2021}, pages 3619--3629.
  Computer Vision Foundation / {IEEE}.

\bibitem[{Stumpf et~al.(2009)Stumpf, Rajaram, Li, Wong, Burnett, Dietterich,
  Sullivan, and Herlocker}]{Stumpf2009}
Simone Stumpf, Vidya Rajaram, Lida Li, Weng-Keen Wong, Margaret Burnett, Thomas
  Dietterich, Erin Sullivan, and Jonathan Herlocker. 2009.
\newblock \href {https://doi.org/10.1016/j.ijhcs.2009.03.004} {Interacting
  meaningfully with machine learning systems: Three experiments}.
\newblock \emph{Int. J. Hum.-Comput. Stud.}, 67(8):639–662.

\bibitem[{Teso et~al.(2021)Teso, Bontempelli, Giunchiglia, and
  Passerini}]{teso2021interactive}
Stefano Teso, Andrea Bontempelli, Fausto Giunchiglia, and Andrea Passerini.
  2021.
\newblock Interactive label cleaning with example-based explanations.
\newblock \emph{Proceedings of the Thirty-fifth Conference on Neural
  Information Processing Systems}.

\bibitem[{Teso and Kersting(2019)}]{teso2019}
Stefano Teso and Kristian Kersting. 2019.
\newblock \href {https://doi.org/10.1145/3306618.3314293} {Explanatory
  interactive machine learning}.
\newblock In \emph{Proceedings of the 2019 AAAI/ACM Conference on AI, Ethics,
  and Society}, AIES '19, page 239–245, New York, NY, USA. Association for
  Computing Machinery.

\bibitem[{Tu et~al.(2020)Tu, Lalwani, Gella, and He}]{tu2020empirical}
Lifu Tu, Garima Lalwani, Spandana Gella, and He~He. 2020.
\newblock An empirical study on robustness to spurious correlations using
  pre-trained language models.
\newblock \emph{Transactions of the Association for Computational Linguistics},
  8:621--633.

\bibitem[{Wang et~al.(2021)Wang, Yang, and Wang}]{wang2021identifying}
Tianlu Wang, Diyi Yang, and Xuezhi Wang. 2021.
\newblock Identifying and mitigating spurious correlations for improving
  robustness in nlp models.
\newblock \emph{arXiv preprint arXiv:2110.07736}.

\bibitem[{Wang and Culotta(2021)}]{wang2021robustness}
Zhao Wang and Aron Culotta. 2021.
\newblock Robustness to spurious correlations in text classification via
  automatically generated counterfactuals.
\newblock In \emph{AAAI}.

\bibitem[{Wei and Zou(2019)}]{wei-zou-2019-eda}
Jason Wei and Kai Zou. 2019.
\newblock \href {https://doi.org/10.18653/v1/D19-1670} {{EDA}: Easy data
  augmentation techniques for boosting performance on text classification
  tasks}.
\newblock In \emph{Proceedings of the 2019 Conference on Empirical Methods in
  Natural Language Processing and the 9th International Joint Conference on
  Natural Language Processing (EMNLP-IJCNLP)}, pages 6382--6388, Hong Kong,
  China. Association for Computational Linguistics.

\bibitem[{Wu et~al.(2021)Wu, Xiao, Sun, Zhang, Ma, and He}]{wu2021survey}
Xingjiao Wu, Luwei Xiao, Yixuan Sun, Junhang Zhang, Tianlong Ma, and Liang He.
  2021.
\newblock A survey of human-in-the-loop for machine learning.
\newblock \emph{arXiv preprint arXiv:2108.00941}.

\bibitem[{Yang et~al.(2020{\natexlab{a}})Yang, Kenny, Ng, Yang, Smyth, and
  Dong}]{yang2020generating}
Linyi Yang, Eoin Kenny, Tin Lok~James Ng, Yi~Yang, Barry Smyth, and Ruihai
  Dong. 2020{\natexlab{a}}.
\newblock Generating plausible counterfactual explanations for deep
  transformers in financial text classification.
\newblock In \emph{Proceedings of the 28th International Conference on
  Computational Linguistics}, pages 6150--6160.

\bibitem[{Yang et~al.(2021)Yang, Li, Cunningham, Zhang, Smyth, and
  Dong}]{yang-etal-2021-exploring}
Linyi Yang, Jiazheng Li, Padraig Cunningham, Yue Zhang, Barry Smyth, and Ruihai
  Dong. 2021.
\newblock \href {https://doi.org/10.18653/v1/2021.acl-long.26} {Exploring the
  efficacy of automatically generated counterfactuals for sentiment analysis}.
\newblock In \emph{Proceedings of the 59th Annual Meeting of the Association
  for Computational Linguistics and the 11th International Joint Conference on
  Natural Language Processing (Volume 1: Long Papers)}, pages 306--316, Online.
  Association for Computational Linguistics.

\bibitem[{Yang et~al.(2020{\natexlab{b}})Yang, Ng, Smyth, and
  Dong}]{yang2020html}
Linyi Yang, Tin Lok~James Ng, Barry Smyth, and Riuhai Dong. 2020{\natexlab{b}}.
\newblock Html: Hierarchical transformer-based multi-task learning for
  volatility prediction.
\newblock In \emph{Proceedings of The Web Conference 2020}, pages 441--451.

\bibitem[{Yao et~al.(2021)Yao, Chen, Ye, Jin, and Ren}]{yao2021refining}
Huihan Yao, Ying Chen, Qinyuan Ye, Xisen Jin, and Xiang Ren. 2021.
\newblock Refining neural networks with compositional explanations.
\newblock \emph{arXiv preprint arXiv:2103.10415}.

\bibitem[{Zaidan et~al.(2007)Zaidan, Eisner, and
  Piatko}]{zaidan-etal-2007-using}
Omar Zaidan, Jason Eisner, and Christine Piatko. 2007.
\newblock \href {https://aclanthology.org/N07-1033} {Using {``}annotator
  rationales{''} to improve machine learning for text categorization}.
\newblock In \emph{Human Language Technologies 2007: The Conference of the
  North {A}merican Chapter of the Association for Computational Linguistics;
  Proceedings of the Main Conference}, pages 260--267, Rochester, New York.
  Association for Computational Linguistics.

\bibitem[{Zhang et~al.(2015)Zhang, Zhao, and LeCun}]{zhang2015character}
Xiang Zhang, Junbo Zhao, and Yann LeCun. 2015.
\newblock Character-level convolutional networks for text classification.
\newblock \emph{Advances in neural information processing systems},
  28:649--657.

\end{thebibliography}
\bibliographystyle{acl_natbib}

\end{document}